\documentclass[10.5]{article}
\date{}
\usepackage{fontspec}
\usepackage[a4paper,margin=0.72in]{geometry}
\usepackage{graphicx}
\usepackage{booktabs}
\usepackage{microtype}
\usepackage{xcolor}
\usepackage{titlesec}
\usepackage{enumitem}
\usepackage{float}
\usepackage{amssymb}
\usepackage{xspace}
\usepackage{amsmath}
\usepackage{booktabs}
\usepackage{subcaption}
\usepackage{wrapfig}
\usepackage{multirow}
\usepackage[most]{tcolorbox}
\usepackage{xcolor}

\usepackage{wrapfig}
\usepackage{tabularx}
\usepackage{makecell}

\definecolor{questionblue}{HTML}{EAF2F8}
\definecolor{questionborder}{HTML}{2874A6}

\titleformat{\section}{\large\bfseries}{\thesection}{0.5em}{}
\titlespacing*{\section}{0pt}{8pt}{3pt}
\setlist[itemize]{leftmargin=1.2em,itemsep=1pt,topsep=2pt}

\def\onedot{.\xspace}
\def\eg{\emph{e.g}\onedot} 
\def\ie{\emph{i.e}\onedot}

\begin{document}

\title{
\textbf{Can Bayesian Optimization Efficiently Find a Strong Single Expert in Neural Thickets?}
}

\author{
  \normalsize
  Nigel Bastian Cendra\textsuperscript{1,*} \quad
  Abdelhamid Ezzerg\textsuperscript{1} \quad
  Fernando Julio Cendra\textsuperscript{2} \quad
  Jeremias Knoblauch\textsuperscript{1} \quad
  Jakob Zeitler\textsuperscript{3,*}
}
\maketitle

\begingroup
\renewcommand{\thefootnote}{}
\footnotetext{
  \textsuperscript{1} University College London \quad
  \textsuperscript{2} Institut Polytechnique de Paris \quad
  \textsuperscript{3} University of Oxford \quad
  \textsuperscript{*} Corresponding authors.
}
\addtocounter{footnote}{-1}
\endgroup
\vspace{-10mm}

\begin{abstract}
\normalsize
\noindent Gradient-free post-training has emerged as a compelling alternative to gradient-based optimization for large language models (LLMs), but existing approaches remain costly. We ask whether structured search can identify a strong single expert under a modest evaluation budget. Motivated by evidence that useful weight updates lie in low-dimensional subspaces, we apply Bayesian optimization within a random linear embedding of weight space. Our method requires no backpropagation and uses a Gaussian process surrogate to guide candidate evaluations efficiently. Across several reasoning benchmarks with Qwen2.5-Instruct models from 0.5B to 3B parameters, Bayesian optimization using five times less candidate evaluations matches or exceeds RandOpt. These results show that surrogate-guided search can substantially reduce the evaluation cost of gradient-free post-training while producing stronger deployable single experts.
\end{abstract}
\section{Introduction}

\begin{wrapfigure}{r}{0.45\textwidth}
    \vspace{-0.5cm}
    \centering
    \includegraphics[width=0.45\textwidth]{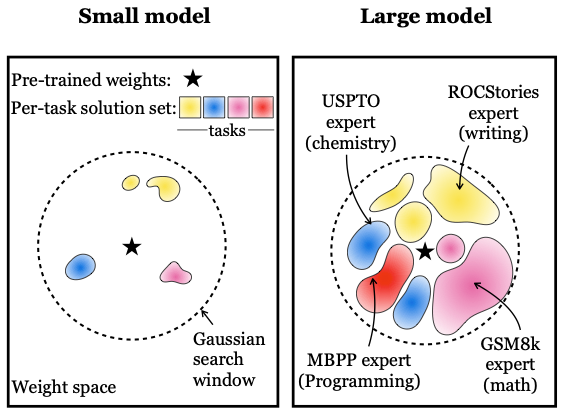}
    \caption{Small models require intelligent search, whereas large models contain many promising solutions for random search. Figure adapted from~\cite{gan2026neuralthickets}.}
    \label{fig:randopt}
    \vspace{-0.4cm}
\end{wrapfigure}
Post-training of large language models is dominated by gradient-based methods such as PPO~\cite{schulman2017ppo} and GRPO~\cite{shao2024deepseekmath}, but a growing line of work shows that \textbf{gradient-free alternatives are competitive at scale}~\cite{qiu2026evolution, sarkar2025evolutionstrategieshyperscale, gan2026neuralthickets}. Neural Thickets~\cite{gan2026neuralthickets} offers a structural explanation: well-pretrained models are surrounded by a dense population of task-specialist weight perturbations (Fig.~\ref{fig:randopt}), therefore sampling random Gaussian perturbations, scoring them on a selection set, keeping the top $K$, and ensembling their predictions by majority vote (RandOpt) can match standard post-training methods without a single backward pass.

A complementary perspective is that useful weight updates concentrate in a subspace far smaller than the parameter count~\cite{li2018measuring, aghajanyan2020intrinsic,hu2022lora}, and when the effective dimension is low, scalar rewards suffice to guide search~\cite{malladi2023finetuning}. How small depends sharply on the training signal: TinyLoRA~\cite{morris2026learningreason13parameters} shows that reward-driven updates need $100$--$1000\times$ fewer parameters than supervised ones, reaching $91\%$ on GSM8K using Qwen2.5 with only $13$ trained parameters. This structure admits two complementary levers: making each candidate cheaper to generate, as in EGGROLL~\cite{sarkar2025evolutionstrategieshyperscale}, whose low-rank candidate perturbations (with full-rank aggregate updates) enable Evolutionary Strategies at population sizes in the hundreds of thousands; or making the search itself smaller, so that far fewer candidates are needed. We take the second, restricting search to a low-dimensional subspace and prioritizing sample efficiency over population-scale efficiency.

Even where random sampling succeeds, however, it is expensive along two axes. (1) \emph{Search cost}: each candidate is scored by generating a full solution for every item in the selection set, so a population of $N$ candidates costs $N \times |\mathcal{D}_{sel}|$ generations, each of which is hundreds of autoregressive decoding steps. RandOpt~\cite{gan2026neuralthickets} uses populations of several thousand: at $N = 5{,}000$ on a $200$-item selection set this is $10^{6}$ generations per run. This cost cannot be escaped by simply sampling less, since the best of $N$ random draws improves slowly in $N$; at modest budgets outcomes become high-variance across runs rather than reliably good. (2) \emph{Deployment cost}: the strongest results rely on majority voting over $K = 50$ perturbed models, so each test query costs $50$ generations and serving requires $K$ distinct weight configurations rather than one. The search cost is paid once; this one is paid on every query, for the lifetime of the model. Falling back to a single model is expensive in accuracy: on Qwen2.5-1.5B-Instruct, RandOpt reports $59.7$ at $K = 50$ against $48.5$ at $K = 1$ MATH-500, and $76.4$ against $67.9$ on GSM8K, drops of $11.2$ and $8.5$ points~\cite{gan2026neuralthickets}. The question is therefore one of efficiency, not feasibility:

\begin{tcolorbox}[
    colback=questionblue,
    colframe=questionborder,
    boxrule=0.8pt,
    arc=2mm,
    left=4mm,
    right=4mm,
    top=1mm,
    bottom=1mm
]
\textbf{Research Question.}
Given a fixed and modest evaluation budget and limited selection data, can a
structured gradient-free method find a stronger single expert than random sampling?
\end{tcolorbox}

Bayesian optimization (BO) is the natural candidate for this setting: it is designed for expensive, noisy, low-dimensional black-box problems, using a Gaussian process (GP) surrogate to decide where to evaluate next, so that each costly evaluation is maximally informative \cite{snoek2012practical}. The surrogate also offers something sampling cannot: a denoised estimate of each candidate's true fitness via the posterior mean, which mitigates the winner's curse inherent in selecting candidates by their raw observed rewards. We therefore run BO within a random linear embedding of weight space~\cite{wang2016bayesian}, keeping the entire pipeline free of backpropagation, and treat the gap between BO and random search at matched budget as a measure of what the surrogate buys. In preliminary experiments on Countdown \cite{gandhi2024streamsearchsoslearning}, GSM8k~\cite{DBLP:journals/corr/abs-2110-14168}, and MATH500~\cite{Mayilvahanan2026unexpected} with Qwen-2.5-Instruct~\cite{qwen2025qwen25technicalreport} 0.5B, 1.5B, and 3B models, BO with 200 candidate evaluations matches or exceeds RandOpt with 1000 at $K=1$, and reaches selection set rewards that random sampling never attains, though only part of this gain transfers: beyond a threshold, further selection reward reflects noise in the selection set rather than better models.

\section{Background}
\noindent \textbf{RandOpt.} Let $\theta_0 \in \mathbb{R}^D$ denote pretrained weights. RandOpt
draws $P$ candidate perturbations $\epsilon_p \sim \mathcal{N}(0, I_D)$, forms
$\theta_p = \theta_0 + \sigma\epsilon_p$, scores each by its reward on a selection set
$\mathcal{D}_{sel}$, and retains the top $K$. At inference the $K$ models each generate an answer
and the majority vote is returned. No gradients are computed at any stage: the entire cost is
$P \times |\mathcal{D}_{sel}|$ generations for search, and $K$ generations per query at deployment.
Neural Thickets~\cite{gan2026neuralthickets} shows this suffices to match gradient-based
post-training for sufficiently large models, because task-specialist perturbations are dense around
$\theta_0$. 

\noindent \textbf{Evolutionary Strategy.} RandOpt shares its sampling step with evolution strategies (ES)~\cite{salimans2017evolutionstrategies}: both draw isotropic Gaussian perturbations of $\theta_0$ and score them. They differ in what they do with the scores, ES aggregates all of them into a gradient estimate and takes a step, while RandOpt selects
the top $K$ and keeps them as models.

\noindent \textbf{Bayesian Optimization.} Bayesian Optimization (BO) targets expensive, noisy black-box objectives. It fits a probabilistic surrogate (\eg Gaussian Process) to the evaluations seen so far and
chooses the next point by maximizing an acquisition function that trades predicted value against
uncertainty; in high dimensions this is typically done within a low-dimensional linear
subspace~\cite{wang2016bayesian}. Two properties matter here. First, each evaluation is chosen using
all previous ones, so the budget is spent where it is most informative, this is the sample
efficiency we are after. Second, the surrogate supplies a posterior mean at every point: an estimate
of true reward that differs from the observed score. Selecting a final candidate by posterior mean
rather than by best observed reward which in principle shrinks the estimates of candidates whose high scores are partly luck.

\section{Method}
\begin{figure}[H]
    \centering
    \includegraphics[width=0.95\linewidth]{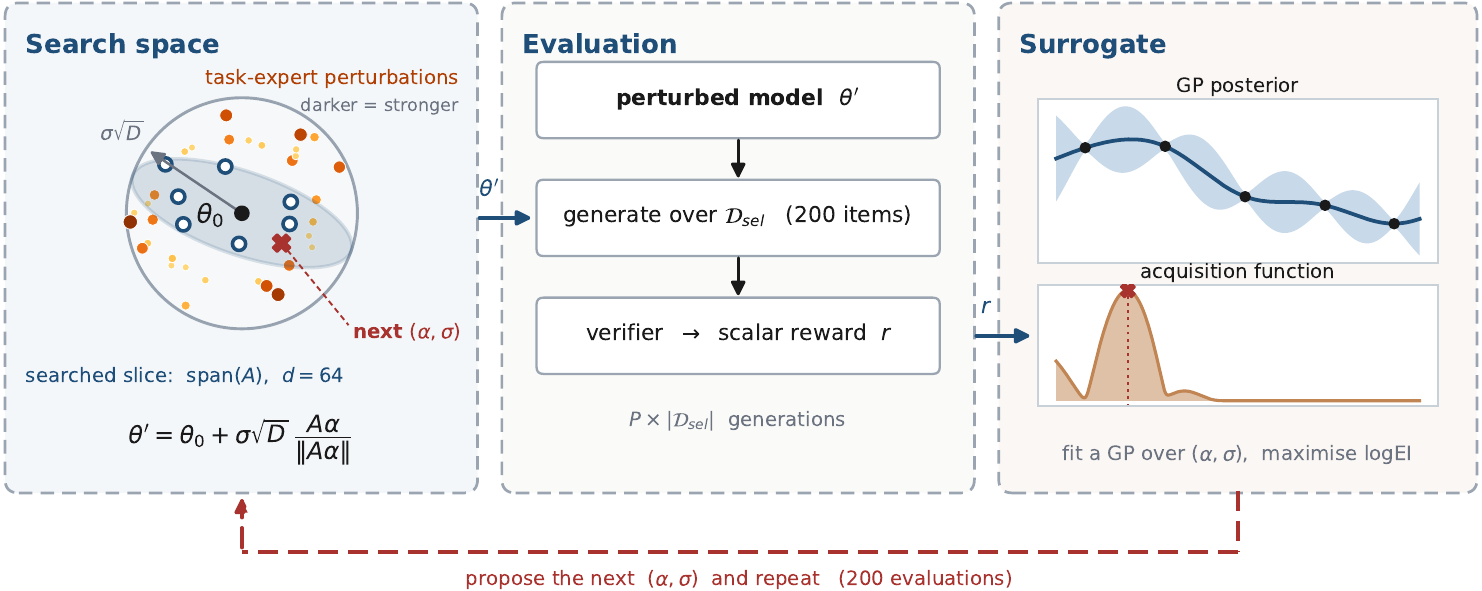}
    \caption{Overview of the method. Open circles are evaluated candidates, confined to $\mathrm{span}(A)$ at distance $\sigma\sqrt{D}$ from $\theta_0$; small dots are task-expert perturbations. The GP is fitted to observed rewards and proposes the next $(\alpha,\sigma)$.}
    \label{fig:method_overview}
\end{figure}

RandOpt samples $\epsilon_p$ isotropically in $\mathbb{R}^D$. We instead search within a random linear embedding: construct $A = [e_1, \dots, e_d] \in \mathbb{R}^{D \times d}$ of $d$ Gaussian directions spans a $d$-dimensional subspace of weight space, and a candidate is specified by its coefficients in that subspace as illustrated in Fig.~\ref{fig:method_overview}. A candidate is parametrized by a coefficient vector $\alpha \in \mathbb{R}^{d}$ and a scale
$\sigma \in \mathbb{R}_{>0}$:
\begin{equation*}
\theta' \;=\; \theta_0 \;+\; \sigma \cdot \sqrt{D} \cdot
\frac{A\alpha}{\lVert A\alpha \rVert}.
\end{equation*}
The perturbation therefore has norm $\|\theta' - \theta_0\| = \sigma\sqrt{D}$, matching the expected norm of a RandOpt perturbation $\sigma\epsilon \sim \mathcal{N}(0, \sigma^{2} I_D)$, for which $\mathbb{E}\|\sigma\epsilon\|^{2} = D\sigma^{2}$. At equal $\sigma$ the two methods move the same distance from $\theta_0$, so our comparisons are matched in perturbation scale as well as in evaluation budget. The normalization decouples the two search variables: $\alpha$ determines only the direction of the perturbation (it is invariant to rescaling of $\alpha$), while $\sigma$ alone sets its magnitude. The search space is therefore $(d{+}1)$-dimensional, with $d = 64$ and $\sigma$ searched on a log scale, regardless of model size $D$.

Each candidate is scored by its reward on a selection set $D_{\mathrm{sel}}$ (accuracy under the task's verifier). We compare three methods under the same evaluation budget $N$: \emph{BO}, which fits a GP surrogate to observed scores and proposes the next $(\alpha, \sigma)$ by maximizing an acquisition function; random search, which draws $(\alpha, \sigma)$ from a fixed prior in the same subspace, isolating the effect of the surrogate; and RandOpt \cite{gan2026neuralthickets}, which samples full-dimensional scalar Gaussian perturbations of $\theta$ as in prior work and serves as an external baseline. The final model is the candidate with the highest observed reward or the highest GP posterior mean, \ie \ the surrogate's estimate of true reward, which should penalizes candidates whose high scores are due to high variability; we compare both selection rules. Held-out test data is evaluated exactly once, by the final selected model, and all methods share prompts, parser, and evaluation stack.

\section{Preliminary Results}

\noindent \textbf{Sample efficiency at $K{=}1$.} With a $5\times$ smaller budget, BO matches or beats RandOpt-1000 at $K{=}1$ in most setting (Tab.~\ref{tab:combined_results}): Countdown-0.5B ($9.34$ vs.\ $6.81$), Countdown-1.5B ($15.05$ vs.\ $14.60$), Countdown-3B ($19.71$ vs.\ $18.89$), GSM8k-1.5B ($67.78$ vs.\ $66.13$) and MATH500-1.5B ($40.43$ vs.\ $39.80$), and is at parity on GSM8k-0.5B ($43.92$ vs.\ $43.99$). The size of the advantage is not arbitrary: it tracks the headroom the neighbourhood offers to any method, measured as the best test accuracy reached by random search or RandOpt minus the base model's. Where the base model is weakest and good perturbations are rare, as on Countdown-0.5B, BO gains most ($+2.53$ over a headroom of $6.46$); where little is available to any method, as on GSM8k-0.5B and MATH500, it gains nothing. BO does not create gains that sampling cannot reach; it extracts available gains at a smaller budget.

\noindent\textbf{MATH500: the sign reverses.} BO attains the highest selection reward at both model sizes ($44.55$ and $63.75$, Tab.~\ref{tab:combined_results}), but test accuracy moves in the opposite direction. At 0.5B it is the lowest of any method ($28.63$), and at both sizes BO falls below the base model ($30.67$ and $41.00$). At 1.5B the two methods with the highest selection reward, BO ($63.75$) and RandOpt-1000 ($63.30$), produce the two lowest test accuracies ($40.43$ and $39.80$), while the two with the lowest selection reward, random search and RandOpt-200 (both $61.80$), produce the two highest ($42.23$ and $41.80$). Neither regime offers much headroom, as on GSM8k-0.5B, where a five point selection gain produced none at test. Where the neighbourhood contains no genuinely better model, selection reward can still be raised by several points, but those gains are fit to $\mathcal{D}_{\mathrm{sel}}$ rather than to the task, and the more effective the search, the more it overfits.

\begin{figure}[H]
    \begin{subfigure}{0.33\linewidth}
        \centering
        \includegraphics[width=\linewidth]{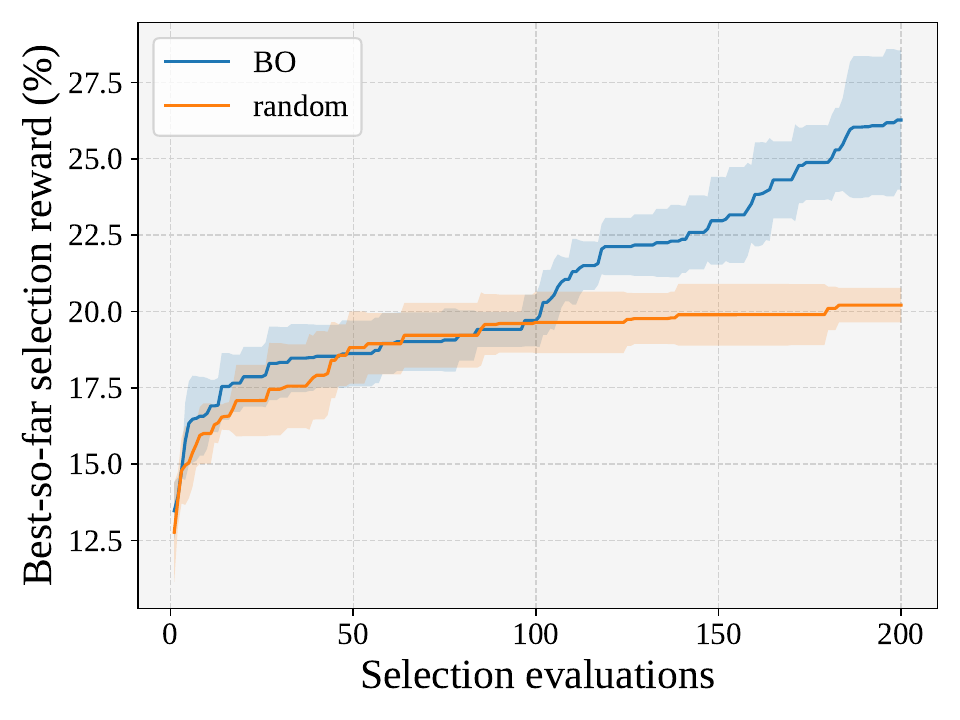}
        \caption{Countdown}
    \end{subfigure}
    \begin{subfigure}{0.33\linewidth}
        \centering
        \includegraphics[width=\linewidth]{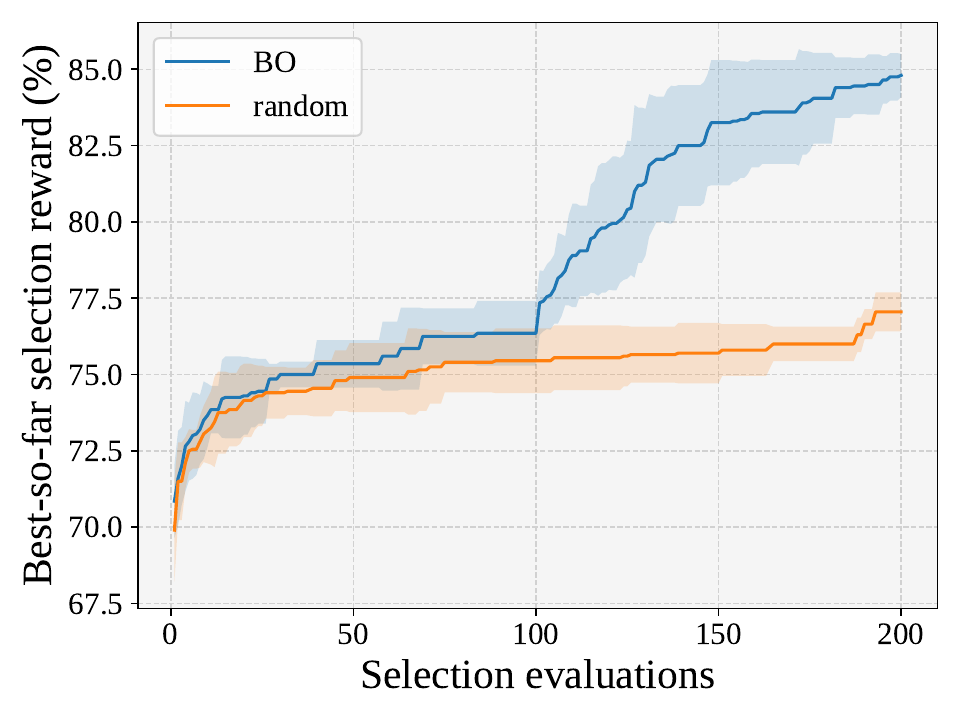}
        \caption{GSM8k}
    \end{subfigure}
    \begin{subfigure}{0.33\linewidth}
        \centering
        \includegraphics[width=\linewidth]{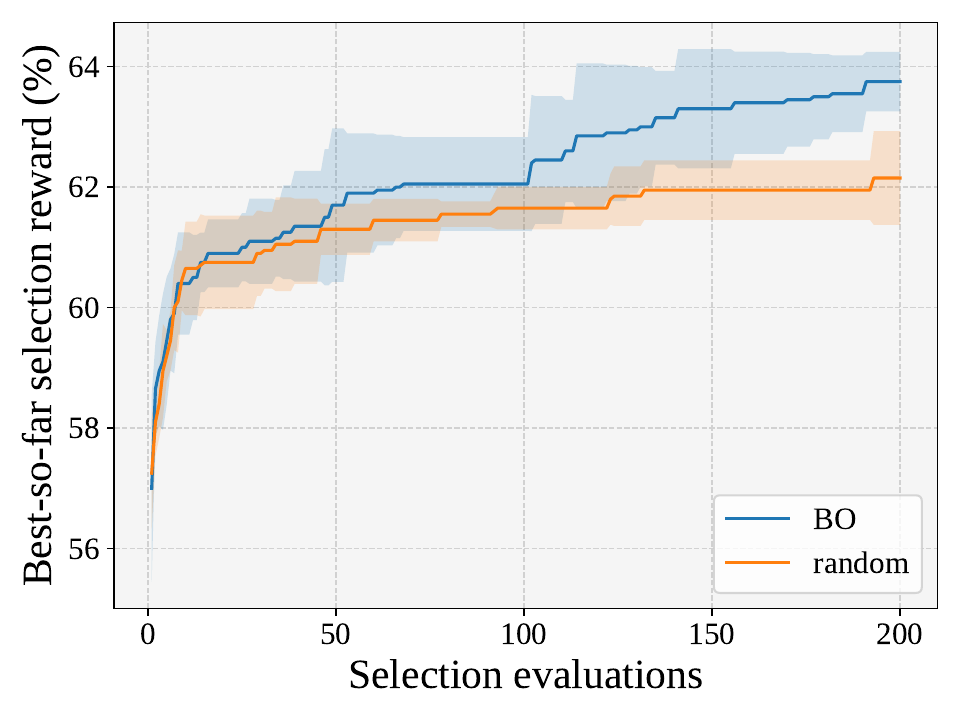}
        \caption{MATH500}
    \end{subfigure}

    \caption{Best-so-far selection rewards for the Qwen2.5-Instruct-1.5B model and its standard deviation.}
    \label{fig:best_so_far_train_rewards}
\end{figure}

\noindent\textbf{Same subspace, different search.} Within a fixed basis, BO and random search differ only in how the next candidate is chosen. BO reaches higher selection reward in all seven settings (Fig.~\ref{fig:best_so_far_train_rewards}), by $1.6$ to $9.97$ points, and higher test accuracy at $K{=}1$ in five of seven; the exceptions are the two MATH500 settings discussed above. The gains we report are therefore attributable to the surrogate rather than to the dimensionality reduction: at matched budget, random search in the basis does not outperform full-dimensional RandOpt sampling, so the low-dimensional parametrization is a concession to tractability rather than a better place to sample. What we can claim is that BO exploits the chosen subspace efficiently. What we cannot claim is how much the choice costs: every optimum we report is by construction contained in a single $64$-dimensional slice, and whether a larger or better-aligned basis raises the ceiling and whether BO's advantage survives there, is untested.

\begin{wrapfigure}{r}{0.45\textwidth}
    \vspace{-0.2cm}
    \centering
    \includegraphics[width=0.45\textwidth]{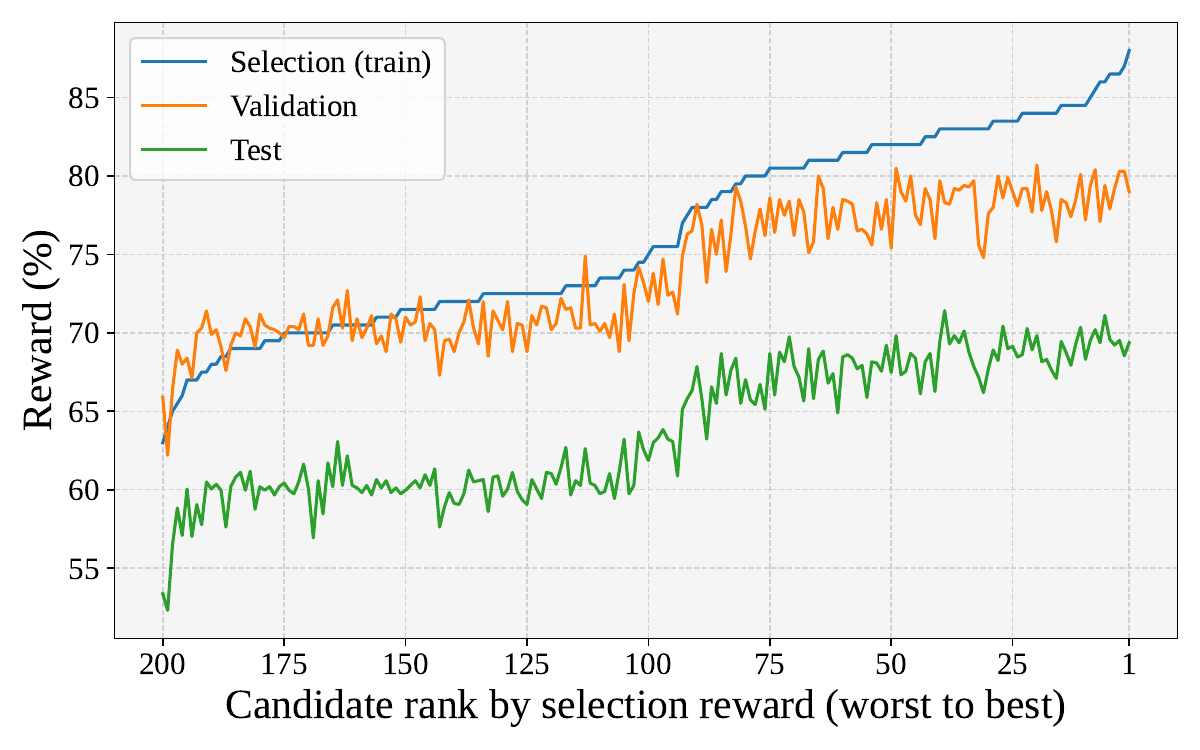}
    \caption{All $200$ candidates from a GSM8k in Qwen2.5-Instruct-1.5B run, ranked by selection reward in descending order, evaluated on the selection set ($200$), validation and test; all three are accuracy.}
    \label{fig:val_test_acc1}
\end{wrapfigure}

\noindent\textbf{Selection reward stops resolving above a threshold.} BO reaches selection set rewards that random sampling never attains (Tab.~\ref{tab:combined_results}), but these gains transfer only in part. Evaluating all $200$ candidates from a GSM8k-1.5B run on a validation set and the test set (Fig.~\ref{fig:val_test_acc1}) shows why. Ranked from worst to best by selection reward, all three curves rise together over the lower three quarters of the range: the selection objective is informative there, and optimising it improves held out accuracy. Above roughly rank $75$, at a selection reward near $80\%$, the curves separate. Selection reward continues climbing to $88\%$ while validation and test accuracy stay flat at approximately $79\%$ and $68\%$. The final eight points of selection reward buy no measurable improvement on either held out set.

Two features identify the cause. Validation and test flatten at the same rank and remain locked together, so the plateau reflects the selection set rather than noise in a single evaluation. And while test sits a constant ten points below validation, reflecting the difficulty difference between the splits, the gap between selection and validation widens from about one point at the worst candidates to eight at the best. Since both are drawn from the same pool, a gap that grows toward the selected extreme is selection bias: the top scores are inflated by favourable draws on $\mathcal{D}_{\mathrm{sel}}$.

\begin{figure}[H]
    \centering

    \begin{subfigure}{0.32\linewidth}
        \centering
        \includegraphics[width=\linewidth]{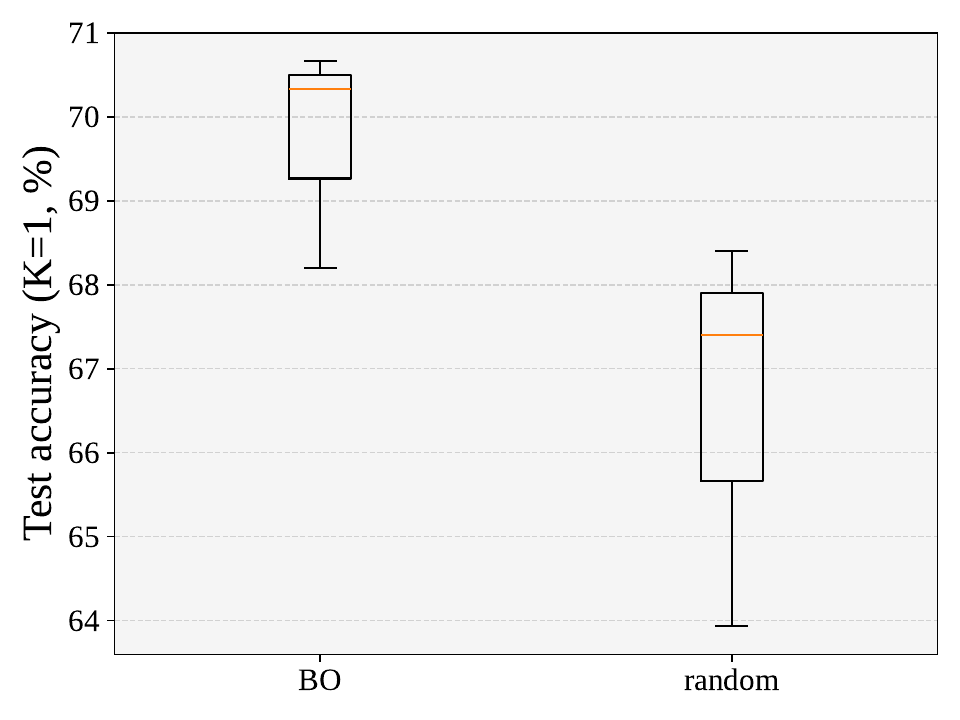}
        \caption{Selection size 200}
    \end{subfigure}
    \hfill
    \begin{subfigure}{0.32\linewidth}
        \centering
        \includegraphics[width=\linewidth]{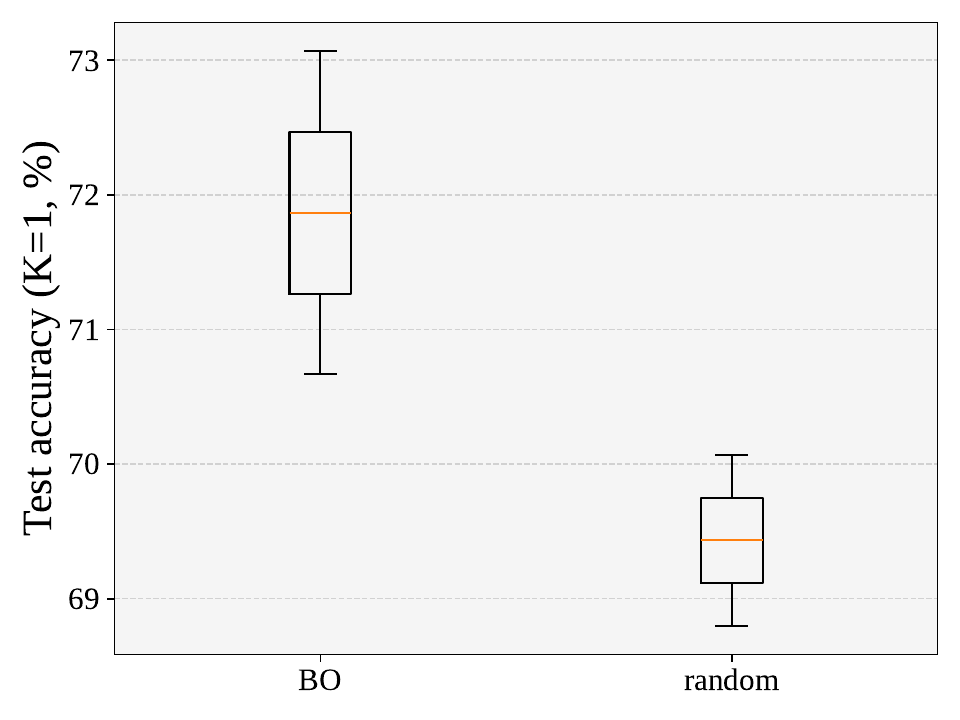}
        \caption{Selection size 1000}
    \end{subfigure}
    \hfill
    \begin{subfigure}{0.32\linewidth}
        \centering
        \includegraphics[width=\linewidth]{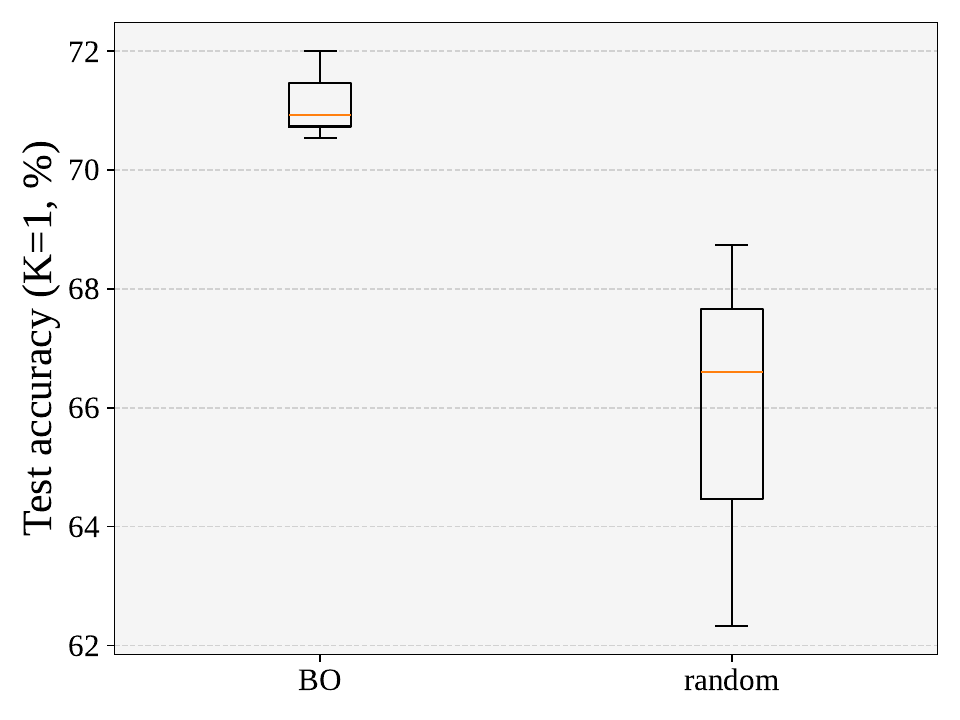}
        \caption{200, qLogNEI~\cite{ament2023unexpected} + posterior mean}
    \end{subfigure}

    \caption{(GSM8K) Comparison of selection strategies and selection sizes. Evaluated on 9 runs and smaller val set.}
    \label{fig:selection_comparison}
\end{figure}

Both interventions act on this quantity. Enlarging $\mathcal{D}_{\mathrm{sel}}$ from $200$ to $1000$ raises the worst run from around $68.2\%$ to $70.6\%$ and the best from $71.0\%$ to $73.5\%$ (Fig.~\ref{fig:selection_comparison}b), and a noise aware variant combining qLogNEI with posterior-mean selection raises the worst of nine runs to $71\%$ at no cost to the best (Fig.~\ref{fig:selection_comparison}c). We do not separate the contributions of the acquisition function and the selection rule, further ablations are required to resolve its contributions.

\noindent \textbf{Majority voting ensembles.} Table~\ref{tab:combined_results} shows that at $K{=}20$ BO matches the alternatives at matched budget on every setting; on Countdown-1.5B its
mean lies well within one standard deviation of both random search and RandOpt-200. The one clear deficit is against RandOpt-1000, which uses five times the search budget, suggesting that ensemble quality continues to benefit from a broader pool of candidates in a way that a concentrated search does not supply. Selecting the $K$ members jointly from the GP, trading individual quality for diversity, is a natural next step.

\begin{table}[ht!]
\centering

\resizebox{0.95\textwidth}{!}{%
\begin{tabular}{cclcccc}
\toprule
Task
& Model size
& Method
& Budget
& Selection reward (\%)
& Test K=1 (\%)
& Test K=20 (\%) \\
\midrule

\multirow{16}{*}{\rotatebox{90}{Countdown}}
& 0.5B
& Base model
& --
& $4.47$
& $0.35$
& -- \\

& 0.5B
& Random search
& 200
& $9.16 \pm 1.31$
& $4.34 \pm 2.35$
& $7.89 \pm 0.81$ \\

& 0.5B
& RandOpt
& 200
& $9.22 \pm 0.84$
& $5.58 \pm 1.58$
& $8.10 \pm 0.22$ \\

& 0.5B
& RandOpt
& 1000
& $13.16 \pm 1.46$
& $6.81 \pm 0.97$
& $10.46 \pm 0.56$ \\

& 0.5B
& \textbf{BO (ours)}
& 200
& $\mathbf{17.12 \pm 1.05}$
& $\mathbf{9.34 \pm 1.15}$
& $\mathbf{11.11 \pm 0.28}$ \\

\cmidrule(lr){2-7}

& 1.5B
& Base model
& --
& $10.68$
& $12.20$
& -- \\

& 1.5B
& Random search
& 200
& $20.21 \pm 0.56$
& $13.16 \pm 1.10$
& $31.61 \pm 1.15$ \\

& 1.5B
& RandOpt
& 200
& $22.58 \pm 2.00$
& $14.03 \pm 1.91$
& $32.05 \pm 1.96$ \\

& 1.5B
& RandOpt
& 1000
& $24.29 \pm 1.27$
& $14.60 \pm 2.24$
& $\mathbf{36.02 \pm 0.88}$ \\

& 1.5B
& \textbf{BO (ours)}
& 200
& $\mathbf{26.27 \pm 2.28}$
& $\mathbf{15.05 \pm 1.86}$
& $31.64 \pm 2.74$ \\

\cmidrule(lr){2-7}

& 3B
& Base model
& --
& $12.53$
& $15.54$
& -- \\

& 3B
& Random search
& 200
& $22.37 \pm 1.02$
& $15.83 \pm 0.69$
& $46.70 \pm 1.41$ \\

& 3B
& RandOpt
& 200
& $23.04 \pm 1.99$
& $16.42 \pm 3.93$
& $46.59 \pm 1.81$ \\

& 3B
& RandOpt
& 1000
& $26.20 \pm 1.05$
& $18.89 \pm 2.93$
& $50.12 \pm 0.54$ \\

& 3B
& \textbf{BO (ours)}
& 200
& $\mathbf{32.34 \pm 3.31}$
& $\mathbf{19.71 \pm 4.24}$
& $\mathbf{50.39 \pm 2.45}$ \\

\midrule

\multirow{10}{*}{\rotatebox{90}{GSM8k}}
& 0.5B
& Base model
& --
& $57.10$
& $43.80$
& -- \\

& 0.5B
& Random search
& 200
& $60.60 \pm 0.71$
& $43.67 \pm 0.43$
& $46.96 \pm 0.41$ \\

& 0.5B
& RandOpt
& 200
& $60.20 \pm 0.76$
& $\mathbf{44.28 \pm 0.97}$
& $46.50 \pm 0.51$ \\

& 0.5B
& RandOpt
& 1000
& $61.40 \pm 0.55$
& $43.99 \pm 0.58$
& $47.13 \pm 0.42$ \\

& 0.5B
& \textbf{BO (ours)}
& 200
& $\mathbf{62.20 \pm 1.27}$
& $43.92 \pm 0.42$
& $\mathbf{47.32 \pm 0.89}$ \\

\cmidrule(lr){2-7}

& 1.5B
& Base model
& --
& $72.00$
& $61.49$
& -- \\

& 1.5B
& Random search
& 200
& $77.05 \pm 0.64$
& $62.49 \pm 1.46$
& $65.34 \pm 0.35$ \\

& 1.5B
& RandOpt
& 200
& $78.70 \pm 1.96$
& $65.84 \pm 1.01$
& $66.82 \pm 1.31$ \\

& 1.5B
& RandOpt
& 1000
& $79.30 \pm 0.57$
& $66.13 \pm 2.43$
& $72.40 \pm 0.82$ \\

& 1.5B
& \textbf{BO (ours)}
& 200
& $\mathbf{84.48 \pm 0.71}$
& $\mathbf{67.78 \pm 2.14}$
& $\mathbf{73.65 \pm 1.66}$ \\

\midrule

\multirow{10}{*}{\rotatebox{90}{MATH500}}
& 0.5B
& Base model
& --
& $37.00$
& $30.67$
& -- \\

& 0.5B
& Random search
& 200
& $42.95 \pm 0.49$
& $\mathbf{30.23 \pm 1.46}$
& $36.63 \pm 0.24$ \\

& 0.5B
& RandOpt
& 200
& $42.80 \pm 0.57$
& $29.47 \pm 1.94$
& $36.47 \pm 1.07$ \\

& 0.5B
& RandOpt
& 1000
& $-$
& $-$
& $-$ \\

& 0.5B
& \textbf{BO (ours)}
& 200
& $\mathbf{44.55 \pm 1.06}$
& $28.63 \pm 1.18$
& $\mathbf{36.70 \pm 1.37}$ \\

\cmidrule(lr){2-7}

& 1.5B
& Base model
& --
& $58.00$
& $41.00$
& -- \\

& 1.5B
& Random search
& 200
& $61.80 \pm 0.45$
& $\mathbf{42.23 \pm 1.18}$
& $52.17 \pm 0.80$ \\

& 1.5B
& RandOpt
& 200
& $61.80 \pm 0.45$
& $41.80 \pm 0.69$
& $52.53 \pm 2.10$ \\

& 1.5B
& RandOpt
& 1000
& $63.30 \pm 0.67$
& $39.80 \pm 3.09$
& $\mathbf{53.60 \pm 1.32}$ \\

& 1.5B
& \textbf{BO (ours)}
& 200
& $\mathbf{63.75 \pm 0.49}$
& $40.43 \pm 1.56$
& $52.00 \pm 2.72$ \\

\bottomrule
\end{tabular}%
}
\caption{
Performance of Qwen2.5-Instruct on the Countdown, GSM8k and MATH500 tasks.
Selection rewards and test accuracies are reported as percentages.
Values are mean $\pm$ standard deviation.
Bold numerical values indicate the highest score within each task and
model-size group. Random search and BO are averaged over $5$ basis seeds $\times$ $2$ search seeds; RandOpt uses no basis and is averaged over $5$ seeds.
}
\label{tab:combined_results}
\end{table}

\begin{wraptable}{r}{0.58\textwidth}
\vspace{-0.5\baselineskip}
\centering

\begin{tabularx}{\linewidth}{@{}Xccc@{}}
\toprule
Method
& \makecell{Selection\\reward (\%)}
& \makecell{Test\\$K{=}1$ (\%)}
& \makecell{Test\\$K{=}20$ (\%)} \\
\midrule
Base model
& 12.53
& 15.54
& -- \\

BO best run
& \textbf{42.18}
& \textbf{34.45}
& \textbf{57.80} \\

BO second-best run
& 32.16
& 21.25
& 55.65 \\

All 10 BO runs
($\pm$ SD)
& $32.34 \pm 3.31$
& $19.71 \pm 4.24$
& $50.39 \pm 2.45$ \\
\bottomrule
\end{tabularx}

\caption{Base model and BO results for Qwen2.5-Instruct-3B across
$5$ basis seeds $\times$ $2$ BO seeds.}
\label{tab:higher_ceiling_res}
\vspace{-0.8\baselineskip}
\end{wraptable}

\noindent\textbf{Evidence of a higher ceiling.} Table~\ref{tab:higher_ceiling_res} shows that in one basis and BO seed combination at Qwen2.5-Instruct-3B, BO located a region with top selection rewards of $42.18\%$, well above the $32.16\%$ reached by the next best run, and its best candidate reached $34.45\%$ test accuracy at $K{=}1$ against $21.25\%$ for that run and $19.71\% \pm 4.24\%$ across all ten. RandOpt never produced a candidate of this quality. Selection reward tracked test accuracy correctly here: a ten point gap in selection reward between the two runs produced a thirteen point gap at test. At $K{=}20$ the same run leads by only $2.15\%$, so the advantage is concentrated in a single candidate rather than spread across the surrounding set. We did not recover this region in later runs, including with the same basis, so the solution is reachable within a subspace BO searches repeatedly but is found only occasionally. This places the difficulty in locating good regions rather than in ranking candidates within one, and
suggests run-to-run variance is dominated by which region a run settles in.


\section{Discussion and Planned Experiments}

\noindent\textbf{Why BO's gains transfer imperfectly.} The selection objective is informative over most of its range and stops being so near the top. Ranking all $200$ candidates of a GSM8k in Qwen2.5-Instruct-1.5B run by selection reward (Fig.~\ref{fig:val_test_acc1}), selection, validation and test accuracy rise together over the lower three quarters, then separate above rank $\sim 75$: selection reward climbs a further eight points while validation and test stay flat. Validation and test flatten at the same rank and remain locked together, so the plateau is a property of the selection set rather than noise in a single evaluation, while the gap between selection and validation, drawn from the same pool, widens toward the selected extreme. BO's gains are therefore real where the signal resolves and inflated where it does not.

\noindent\textbf{Planned experiments.} (i) Scaling to Qwen2.5-Instruct-3B with a larger budget ($N = 500$ rather than $200$). Raising the population increases selection reward in every setting we have run, but improves test accuracy only where the neighbourhood still offers headroom, so the value of additional search budget is not constant across model scales. Full results are pending. (ii) A selection-data sweep ($|\mathcal{D}_{\mathrm{sel}}| = 200$ to $1{,}000$) with the test set fixed, to test whether the saturation threshold moves with resolution as predicted.



{
\bibliographystyle{unsrt}
\bibliography{references}
}

\end{document}